\documentclass{article}

\PassOptionsToPackage{numbers, compress}{natbib}
\usepackage[final]{nips_2017}

\usepackage[utf8]{inputenc} 
\usepackage[T1]{fontenc}    
\usepackage{hyperref}       
\usepackage{url}            
\usepackage{booktabs}       
\usepackage{amsfonts}       
\usepackage{nicefrac}       
\usepackage{microtype}      
\usepackage{amsmath}
\usepackage[pdftex]{graphicx}
\usepackage{algorithm}
\usepackage[noend]{algpseudocode}
\usepackage{url}

\renewcommand{\vec}[1]{\mathbf{#1}}
\let\oldhat\hat
\renewcommand{\hat}[1]{\oldhat{\mathbf{#1}}}

\title{Spline-Based Probability Calibration}

\author{
  Brian Lucena \\
  Lucena Consulting\\
  \texttt{brian@lucenaconsulting.com} \\
 }

\begin{document}

\maketitle

\begin{abstract}
In many classification problems it is desirable to output  {\em well-calibrated} probabilities on the different classes.  We propose a robust, non-parametric method of calibrating probabilities called SplineCalib that utilizes smoothing splines to determine a calibration function. We demonstrate how applying certain transformations as part of the calibration process can improve performance on problems in deep learning and other domains where the scores tend to be "overconfident".  We adapt the approach to multi-class problems and find that better calibration can improve accuracy as well as log-loss by better resolving uncertain cases.  Finally, we present a cross-validated approach to calibration which conserves data. Significant improvements to log-loss and accuracy are shown on several different problems.  We also introduce the \texttt{ml-insights} python package which contains an implementation of the SplineCalib algorithm.
\end{abstract}

\section{Introduction and Motivation}
When building a classification model, it is useful to keep in mind that there are three different variants of classification problems.  In {\em hard classification} the goal is to predict a single category for each instance.  The quality of the classifier is them measured by accuracy, precision, recall, f1-score, or other metrics in a similar vein.  Other situations (primarily in binary classification) call for a {\em ranking classifier}.  Here the goal is to assign a score to each instance such that a higher score indicates higher likelihood of being "in-class".  If desired, (in the binary case) a threshold can be chosen to convert the ranking classifier into a hard classifier.  Alternatively, the performance can be judged by metrics such as the AUROC or a Precision-Recall Curve.  This is akin to measuring performance over the range of hard classifiers obtained by varying the threshold.  In these first two problems, we are concerned with the {\em discriminative} power of the algorithm, i.e. its ability to separate the different classes. A third task is {\em probability prediction}.  This refers to assigning a specific probability distribution across the classes for each instance.  Of course, these probabilities can be considered mere scores to convert this output into a ranking classifier or hard classifier.  However, in this task we typically use metrics such as the log-loss (sometimes referred to as {\em cross-entropy}) or Brier score~\cite{brier1950verification} to measure performance.  These metrics consider not only the discriminative power, but the {\em calibration} of the outputted distributions.  In fact, the Brier score permits an explicit decomposition into a discriminative component (usually called {\em refinement}) and a calibration component.~\cite{murphy1972scalarA}~\cite{murphy1972scalarB}~\cite{murphy1973new}

Calibration is particularly important when the optimal decision to be made is sensitive to the probability of being correct {\em on that particular case}.  While metrics like precision may measure the probability of the entire set of instances which exceed a threshold, they do not account for gradations of probability within that set.  For example, if you set a threshold to achieve a precision of $p$, it is quite likely that that an instance that barely clears the threshold has a probability less than $p$ of being "in-class", while the instances that far exceed the threshold have probability greater than $p$.

In the binary case, the process of {\em probability calibration} refers to taking a set of {\em scores} (which may or may not already purport to be probabilities) and providing a mapping of those scores to the interval $[0,1]$ such that the outputs "behave well" as probabilities.  In other words, when we look at the set of events where the predicted probability was .7, we should expect that, in the long run, 70\% of those events indeed happened.  Both the log-loss and the Brier score have the property that they are optimized when the "true probability" is predicted.  The multi-class case is analogous, except the input and output are vectors of probabilities that sum to one.

One powerful aspect of calibration is that it can be used to improve nearly any existing classification algorithm.  All that is required are a set of scores and truth values to fit the calibration function.  Then the calibration function can be applied to the predictions of the model (in a post-hoc fashion) to yield more accurate predictions.  Although the calibration function is designed to improve metrics like log-loss and the Brier score, we will see that it has the capability to improve accuracy as well.

\section{Previous Work}
There exist two main methods for calibrating scores in a binary classification setting: Platt scaling~\cite{platt1999probabilistic}~\cite{niculescu2005predicting} and Isotonic Regression~\cite{zadrozny2001obtaining}~\cite{zadrozny2002transforming}.  Platt scaling fits a one-variable logistic regression model to predict the outcomes from the scores.  This approach was motivated by work demonstrating that the relationship between the scores given by SVMs and the actual probabilities are often fit well by the sigmoid function.  As one would expect from a strict parametric approach, it works well when the data actually fit the model, but gives poor results when the desired calibration function is not well-approximated by a sigmoid function.

Isotonic Regression~\cite{zadrozny2001obtaining}~\cite{zadrozny2002transforming} fits a piecewise constant calibration function.  This works reasonably well on a wider variety of problems due to its non-parametric approach.  However, piecewise-constant approximations yield room for improvement due to their coarseness.  Another limitation of these approaches is that they do not extend directly to a multi-class setting.  Rather,~\cite{zadrozny2001obtaining} and~\cite{zadrozny2002transforming} propose an indirect approach of fitting multiple binary classifiers and then calibrating and combining them.

The issue of calibration has taken on a particular nuance when using Deep Neural Networks, which tend to be "overconfident".  Thus, models which perform well in terms of accuracy may do poorly in terms of log-loss.  This issue has arisen in Kaggle competitions, which increasingly use log-loss as the competitive metric for image classification problems.  These competitions are primarily on multi-class problems, where the techniques of~\cite{zadrozny2001obtaining} and~\cite{zadrozny2002transforming} are infeasible.  Consequently, many participants resort to the practice of "clipping".  Simply put, clipping refers to assigning a minimum probability $p_{min}$ to every class, and then re-normalizing the probability vector to sum to one.

\section{Algorithm Description}
In a binary classification problem, the scores are one-dimensional, and represent a measure of how likely that instance is to be "in-class" versus "out-of-class".  For multi-class classification, the score is a vector with a value for each possible class.  For simplicity, we assume throughout this paper that all scores are (uncalibrated) probabilities.  For the binary case, this means the scores are in the interval $[0,1]$ where higher numbers mean greater likelihood of being "in-class". For the multi-class scenario with $m$ classes, this means the model outputs a {\em score vector} for each instance, such the score vector has length $m$, the individual values are in $[0,1]$ and the vector sums to one.  Most standard implementations of machine learning algorithms give outputs in these forms, and it is straightforward to re-normalize them if they are not.

\subsection{The {\em SplineCalib} Algorithm}
This paper introduces an algorithm called {\em SplineCalib} which uses smoothing splines to calibrate probabilities.  This spline-based approach is non-parametric like isotonic regression.  However, it typically outperforms isotonic regression, since it has the freedom to fit a cubic spline rather than just a piecewise constant function.  This paper, and the {\em SplineCalib} algorithm, make the following contributions and improvements to probability calibration:
\begin{itemize}
\item Using cubic splines to fit the calibration function, rather than a piecewise constant or sigmoid function.
\item Proposing a transform to be done on the scores prior to calibration, which improves performance for "over-confident" classifiers (such as Naive Bayes and Deep Learning models).
\item Employing a novel approach to multi-class classification that works directly with the results of the multi-class classifier.
\end{itemize}

\subsection{Background on Splines}
Splines arise in the context of fitting non-linear curves to data. (For general references on splines, see~\cite{wahba1990spline}~\cite{hastie2001elements}~\cite{wakefield2013bayesian}). Consider predicting a univariate response $y$ from a univariate predictor $x$.  Suppose we wish to allow a response function that is more complicated that a simple line, but still continuous and relatively smooth, using mean-squared error on an independent test set as our criterion of success.  Thus, we want a curve which fits the training data well (in terms of a least squares error) but not to the point that we have a very wiggly, unrealistic curve with low predictive power.

Splines can be thought of as a natural continuation of the trajectory from linear regression to polynomial regression.  In polynomial regression, higher order terms such as $x^2$ and $x^3$ are added to the set of variables on which the regression is performed. However, this approach is often ineffective since the "right" polynomial should be locally defined.  Spline methods expand the basis (i.e. the set of predictors derived from the value $x$) beyond mere terms of the form $x^k$.  Rather, the basis is chosen with respect to a set of knots, and done in a way such that the resulting function (obtained by multiplying the basis functions by the fitted coefficients) has "nice" properties such as continuity and smoothness.  

Regression splines use a fixed set of knots such that a different polynomial is used in each interval.  The more knots are used, the better the fit to training data, but the higher the risk of overfitting.  Smoothing splines~\cite{wahba1990spline} avoid the knot selection problem by using all $x$-values as knots and regularizing via a penalty on the integrated second derivative.  While most of the literature on splines focuses on a least-squares criterion, as in linear regression, in most cases it is straightforward to extend them a log-likelihood criterion (as in logistic regression).  In that case, we minimize the loss function given by:

\begin{equation}
 J(f) = -\sum_{i=1}^n \left[ y_i \log(f(x_i)) + (1-y_i) \log(1-f(x_i))\right] + \frac{1}{2} \lambda \int f^{\prime\prime} (t) dt
 \end{equation}
 
 for a given set of $ {(x_i, y_i)}$. (This is sometimes referred to as {\em non-parametric logistic regression}~\cite{hastie2001elements}). Here, $\lambda$ is a parameter which determines how much to penalize the curvature of the function $f$.  As $\lambda \rightarrow \infty$ no curvature is tolerated, while as $\lambda \rightarrow 0$, one typically gets a very wiggly function which overfits the training data.  Cross-validation can be used to find an appropriate value of $\lambda$.

Of particular interest in this paper is the {\em natural cubic spline basis}.  Given a set of knots $\{\phi_1, \phi_2, \ldots,\phi_K\}$ (with $\phi_i < \phi_j\text{ when }i<j)$, the natural cubic spline basis is given by:
\begin{gather*}
 N_1(x) = 1, \: N_2(x) =x,\: N_{k+2}(x) = d_k(x)-d_{K-1}(x)\\
  \text {for $k$ in $1,2,\ldots,K-2$}\\
\text{where }d_k(x) = \frac{(x-\phi_k)_{+}^3 - (x-\phi_K)_{+}^3}{\phi_K - \phi_k}
\end{gather*}
Note that the size of the resulting basis is equal to the size of the set of knots.

\subsection{Preliminaries}\label{prelim}
Standard approaches to calibration use the typical supervised learning scenario, except with the addition of a third independent {\em calibration set} to go along with the training and test sets.  A model is fit to the training data and is then applied to the features of the calibration set to obtain predictions ${\hat{y}}_{ca}$.  A {\em calibration method} is employed which takes as input ${\hat{y}}_{ca}$ and the corresponding truth values $y_{ca}$ and outputting a calibration function $f$. This function $f$ is then applied to the results of the model to yield calibrated predictions.  We can then compare the results of the calibrated and uncalibrated models on the test set, using log-loss or Brier score as our metric.

We will present a series of algorithms in the paper of increasing complexity, each one building on the previous versions.  Each algorithm will be demonstrated on a particular classification problem.

\section{Binary Classification}
To begin, we present the most basic version of SplineCalib.  Put succinctly, we perform a non-parametric logistic regression on the pair ${\hat{y}}_{ca}$, $y_{ca}$ as defined in Section~\ref{prelim}. Some care must be taken in choosing a set of knots, a spline basis, and the parameter $\lambda$. The detailed method is given by Algorithm~\ref{alg1}.

\begin{algorithm}
\caption{{\em SplineCalib} Basic}\label{alg1}
\begin{algorithmic}[0]
\State {\bf Input:} Vectors $\hat{\vec{y}}$ of predicted probabilities and $\vec{y}$ of corresponding actual $[0,1]$ values
\State {\bf Output:} A calibration function $f: [0,1] \rightarrow [0,1]$\\
\State (1) Sample a set of knots of size $k$ from the unique values of  $\hat{\vec{y}}$  (One could use all the knots, but a sample of size 200 is usually sufficient)

\State (2) Let $\vec{X}$ be the natural basis expansion of the values in $\hat{\vec{y}}$ using the defined set of knots.  ($\vec{X}$ will be a matrix of size $n \times k$)

\State (3) Do a cross-validated $L_2$-regularized Logistic Regression on the pair $(\vec{X},\vec{y}) $ over a range of possible $\lambda$ values. Choose $\lambda^*$ as the value which gives the best cross-validated log-loss.

\State (4) Refit a final $L_2$-regularized Logistic Regression on the pair $(\vec{X},\vec{y})$ using the value $\lambda^*$ found in the previous step.

\State (5) Return the calibration function $f(z): [0,1] \rightarrow [0,1]$ defined by composing the operations of (a) computing the natural basis expansion of $z$ and (b) using the Logistic Regression model from step (4) to predict a probability.

\end{algorithmic}
\end{algorithm}

\subsection{First Binary Classification Example}
Our first example uses data derived from the MIMIC III database~\cite{johnson2016mimic}.  The predictors were 51 features representing the measurements of various lab values and vital signs taken over the first 24 hours of the ICU stay of a patient.  Examples of features include the maximum white blood cell count or the minimum respiration rate (across the first 24 hours). The outcome variable was in-hospital mortality.  The full dataset contained 59,726 hospital stays, which were randomly split: 60\% for training and 20\% each for calibration and testing.

\begin{figure}[b]
  \centering
  \includegraphics[width=1\linewidth]{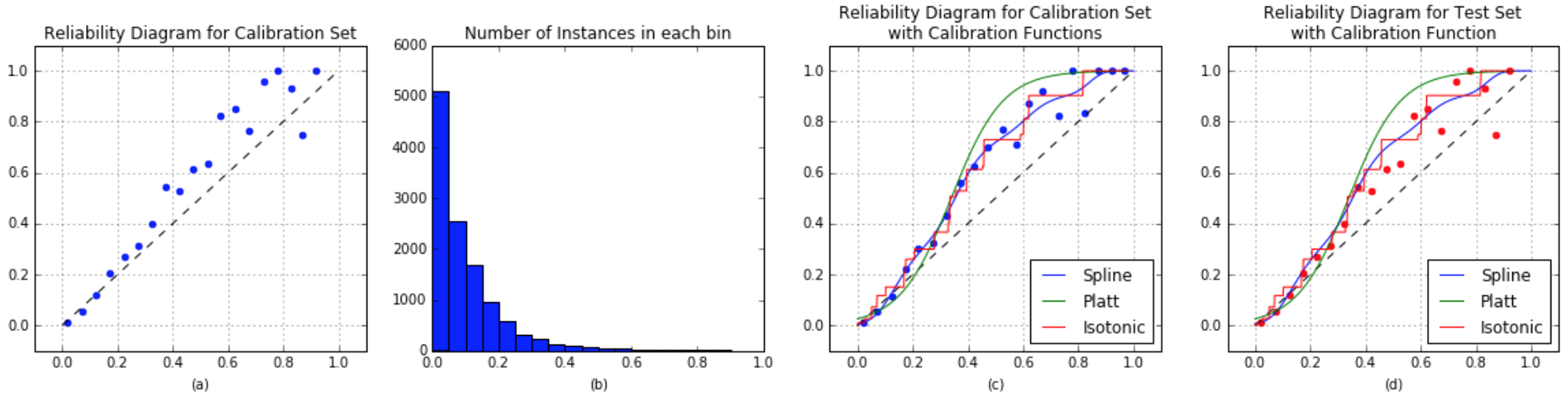}
  \label{rel_calib}
\end{figure}

To begin, we built a standard Random Forest (RF) model with 1000 trees on the training set. We then compute the score (i.e. the proportion of trees which voted "yes") for each instance in the calibration set.  In Figure~\ref{rel_calib}(a), we plot a {\em reliability diagram}~\cite{degroot1983comparison} to assess how well-calibrated the vote percentage is.  A well-calibrated set of scores would have all points at or near the diagonal.  The results indicate that the scores are not well-calibrated.  For example, instances which get a score of around 0.6 tend to be mortalities with probability about 0.8.  Figure~\ref{rel_calib}(b) gives a histogram of the scores, indicating that the smaller scores are much more common (and therefore will factor more heavily in the log-loss calculation).

We use then use the calibration set to determine a calibration function using 3 different methods: Platt scaling, isotonic regression, and  SplineCalib.  Next we assess the log-loss of the original RF model on the test set under the 4 different calibration conditions.  The results are in Table~\ref{LL_RF_NB}.  Spline Calibration greatly outperforms the other variants.  In Figure~\ref{rel_calib} (c) and (d) we see the resulting calibration functions overlaid on the reliability diagrams for the calibration set and test set respectively.  Platt scaling has relatively few degrees of freedom, and so fits the smaller scores but misses badly between 0.4 and 0.8.  Isotonic Regression misses on the smaller scores, has an obvious "staircase" pattern, and overfits in a few places.  SplineCalib effectively "smoothes" the staircase, fits better than Isotonic Regression on the smaller scores, and thus generalizes better to the test set.

\begin{table}[t]
  \caption{Log-loss for MIMIC Random Forest  and Adult Naive Bayes Models}
  \label{LL_RF_NB}
  \centering
  \begin{tabular}{llll}
    \toprule
    &MIMIC-RF &Adult-NB & \\
    \cmidrule{1-4}
    Method     & Log-Loss & Log-Loss      \\
    \midrule
     Uncalibrated & 0.2525&0.7448    \\
    Isotonic Reg.  & 0.2592    & 0.3976     \\
    Platt Scaling  & 0.2535 & 0.4287 \\
    SplineCalib-untransformed & {\bf 0.2442}     & 0.4032    \\
    SplineCalib-compact logit &NA    & {\bf 0.3934 }   \\

    \bottomrule
  \end{tabular}
\end{table}

\section{Calibration for Overconfident Models}
Certain machine learning techniques such as Naive Bayes, and more recently, Deep Neural Networks are known for being {\em overconfident}.  That is, while they may have good (or even excellent) discriminative power, the scores given, when treated as probabilities, tend to be closer to 1 and 0 then they should be.  For example, a Naive Bayes model may give an outcome a predicted probability of .99 which in reality, will only happen 75\% of the time.  If the actual probability changes significantly from 0.99 to 0.999 to 0.9999, this may require a sharp curve in the calibration function, which would be highly penalized by the smoothing spline fitting function.  Thus, it seems reasonable to expect that rescaling the data prior to spline calibration may yield benefits.

\subsection{Compact Logit Function}
To accommodate this, we propose a function from $[0,1] \rightarrow [0,1]$ called the {\em compact logit} function defined by:

$
G_\epsilon(x) = 
\begin{cases}
\frac{(1-2\epsilon))}{(2*\log((1-\epsilon)/\epsilon))}\log\left(\frac{x}{(1-x)}\right)+\frac{1}{2} , & \text{if } x \in [\epsilon, 1-\epsilon] \\
x, 	& \text{if } x \in [0,\epsilon] \cup [1-\epsilon,1] 
\end{cases}
$
\begin{figure}[b]
  \centering
  \includegraphics[width=.8\linewidth]{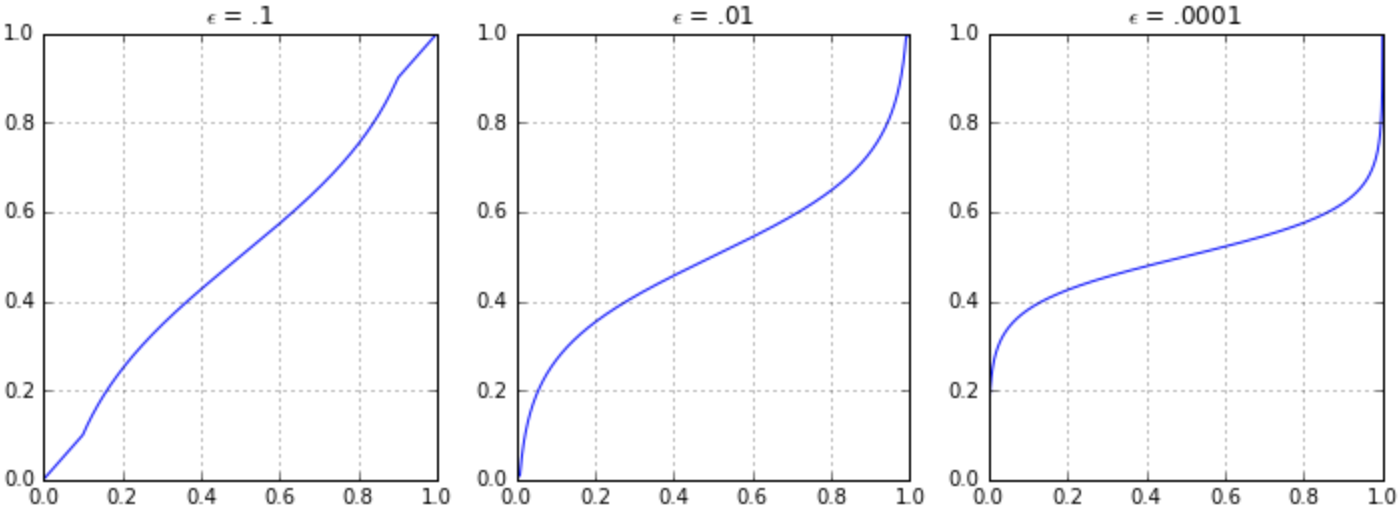}
  \caption{The compact logit function for $\epsilon = .1, \epsilon =.01$, and $ \epsilon =.0001$.}
  \label{compact_logit}
\end{figure}

A plot of this function for a few values of $\epsilon$ is shown in Figure~\ref{compact_logit}.  This primary part of this function is essentially a logit function, truncated, scaled and shifted to have domain and range $[\epsilon, 1-\epsilon]$.  The second case simply makes it a continuous function with domain and range $[0,1]$.  It is easy to verify that $G_\epsilon(\epsilon) = \epsilon$, $G_\epsilon(1-\epsilon) = 1-\epsilon$, and $G_\epsilon(\frac{1}{2}) = \frac{1}{2}$.  The idea behind its usage is to pick a value of $\epsilon$ such that there is no interesting "signal" in the intervals $[0,\epsilon]$ and $[1-\epsilon,1]$.  In this paper, we use the heuristic of choosing $\epsilon = 10^{r-1}$ where $r = \lfloor \log_{10}( \min(1-p_i))\rfloor$ where the $p_i$ are the uncalibrated probabilities (ignoring any probabilities that are exactly 1).

\subsection{Adult Dataset with Naive Bayes}
To illustrate the power of this rescaling, we used the Adult Dataset, also known as the Census Income dataset, available from the UCI Machine Learning Repository~\cite{Lichman:2013}.  This dataset has 32,561 training samples and 16,281 test samples.  The goal is to predict whether or not the individual has income greater than \$50,000/year.  We built a Naive Bayes model using the variables work-class, education-num, marital-status, relationship, race, and sex.

\begin{algorithm}[t]
\caption{{\em SplineCalib} with Compact Logit Transform}\label{alg2}
\begin{algorithmic}[0]
\State {\bf Input:} Vectors $\hat{\vec{y}}$ of predicted probabilities and $\vec{y}$  of actual $[0,1]$ values (from an independent calibration set)
\State {\bf Output:} A calibration function $f: [0,1] \rightarrow [0,1]$\\

\State (1) Let $\vec{y}^\prime = G_\epsilon(\hat{\vec{y}})$ (with abuse of notation to mean the element-wise application of the function)

\State (2) Apply Algorithm~\ref{alg1} on $\vec{y}^\prime, \vec{y}$ to get a function $f_1(x)$

\State (3) Return the function $f_2(x) = f_1(G_\epsilon(x))$

\end{algorithmic}
\end{algorithm}

We repeated the same procedure as in the previous example, although this time we used SplineCalib in two variants.  The first variant had no scaling, while in the second variant we scaled the scores through the compact logit function.  The results are shown in Table~\ref{LL_RF_NB}.  As expected, the uncalibrated version does very poorly due to the overconfidence of the Naive Bayes approach.  Isotonic regression does better than the untransformed SplineCalib but not as well as the transformed version.  This makes sense since Isotonic Regression relies only on the order of the scores and so is insensitive to these scaling issues.  Platt scaling performs the worst of the calibrated versions, though is still a huge improvement over the uncalibrated methods.

\section{Multi-class Extension}
The issue of extending binary calibrators to the multi-class setting had been visited before~\cite{zadrozny2001obtaining}~\cite{zadrozny2002transforming}.  The smoothing spline has a natural generalization to higher dimensions known as {\em thin-plate splines}. However, the number of knots needed to maintain the same resolution would grow exponentially with the number of classes, making a direct generalization infeasible.  Instead, we propose a different approach which leverages the binary classifier while still working directly with the multi-class output.  Put succinctly, we calibrate each column separately and then normalize the results so that the predicted probabilities sum to 1 for each instance.  Essentially, this makes the assumption that the calibration of a particular class is independent of how the remaining probability is distributed amongst the other classes. However, it works well in practice, as we will see.  The precise method is given by Algorithm~\ref{alg3}.  We use $\mathcal{S}^m$ to denote the {\em probability simplex} in $m$ dimensions (i.e. the set of vectors $\vec{y} \in [0,1]^n$ such that $\sum_i y_i = 1$.)

\begin{algorithm}
\caption{SplineCalib for Multi-class Calibration}\label{alg3}
\begin{algorithmic}[0]
\State {\bf Input:} A $n \times m$  matrix $\hat{\vec{Y}}$ of predicted probabilities for each class and a vector $\vec{y}$  of actual $[1, \ldots, m]$ class values. 
\State {\bf Output:} A multiclass calibration function $f: \mathcal{S}^m \rightarrow \mathcal{S}^m$\\

\State (1) Let $\hat{\vec{y_i}}$ be the $i$th column of $\hat{\vec{Y}}$ and let ${\vec{y_i}}$ be given by $I_{i}(\vec{y})$ (the indicator function applied element-wise) for $i$ in $1, \ldots,m$.

\State (2) Use Algorithm~\ref{alg1} or Algorithm~\ref{alg2} on each pair $(\hat{\vec{y_i}},{\vec{y_i}})$ to obtain a function $f_i$

\State (3) Return $f: \mathcal{S}^m \rightarrow \mathcal{S}^m$ where $f$ is given by $f(\vec{x}) = \frac{f_i(x_i)}{\sum_j f_j(x_j)}$

\end{algorithmic}
\end{algorithm}

\subsection{CIFAR10 and Convolutional Neural Nets}
\begin{figure}[b]
  \centering
  \includegraphics[width=1\linewidth]{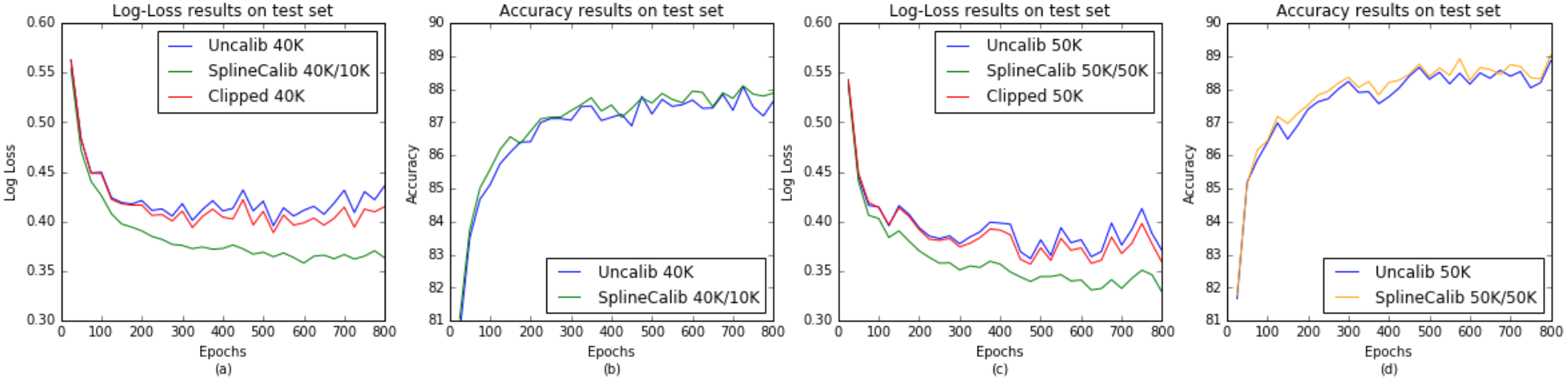}
  \caption{Results on CIFAR-10.}
  \label{CIFAR_10_1}
\end{figure}

To demonstrate SplineCalib in a multi-class setting, we used the CIFAR-10 image dataset~\cite{krizhevsky2009learning}.  This dataset has a designated training set of 50,000 images and a test set of 10,000 images, each equally distributed amongst 10 classes.  We used a Convolutional Neural Network (CNN) model from the file "cifar10\_cnn.py" on the Examples page of the Keras github repository~\cite{keras_example}.  We set aside 10,000 of the 50,000 training images as our calibration set, and trained the CNN on the remaining 40,000 images.  We ran the model for 800 epochs.  After every 25 epochs, we computed the log-loss of the test set under three conditions: uncalibrated, clipped, and calibrated (using Algorithm~\ref{alg3} on the calibration set).  For the clipped version, we took the best result out of 4 different clipping parameters $p_{min} = (0.01,0.001,0.0001, 0.00001)$.  We also computed the accuracy for both the uncalibrated and calibrated models.  The results are in Figure~\ref{CIFAR_10_1} (a) and (b).   After 800 epochs, the uncalibrated log-loss was 0.4361, with clipping it was 0.4150 and with SplineCalib calibration it was 0.3633.  The uncalibrated accuracy was 87.64\% while the calibrated accuracy was 87.88\%.

There are several points worth making about Figure~\ref{CIFAR_10_1}.  The log-loss is dramatically lower with SplineCalib calibration than without and the magnitude of log-loss improvement achieved by SplineCalib far exceeds that obtained by clipping.  Moreover, the log-loss trajectory is much smoother and consistently downward than the uncalibrated and clipped versions.  One potential explanation is that is that we are significantly reducing the {\em calibration} component of the error and therefore left only with the {\em refinement} component.

With respect to accuracy, we see that the calibrated version again outperforms.  Though the magnitude of improvement is less drastic, it shows that better calibration can improve accuracy, presumably due to making better judgments on the borderline cases.

\section{Cross-Validated Calibration}
In this section, we address one of the major limitations of calibration approaches - the need for an independent calibration set.  When training data is scarce, it may not be clear if it is "worth it" to set aside data for calibration.  In terms of the previous example, it may be true that the model trained on 40K examples and calibrated on 10K examples outperforms the uncalibrated model trained on 40K examples.  But does it outperform the uncalibrated model trained on 50K examples?  In general, how can one choose the "right" size of a calibration set so that the gain from calibration is not offset by the loss incurred by having less training data for the model?

The final variant of {\em SplineCalib} described in Algorithm~\ref{alg4} overcomes this problem.  By generating scores for calibration in a cross-validated manner, and training the model on the full training set, we effectively use {\em all} of the data for both model training and calibration.  One caveat is that the scores used for calibration come from slightly different models, each of which will, in turn, be slightly different from the final model to which the calibration is applied in practice.  These differences can be lessened by choosing a larger number of folds, at the cost of increasing the time required for computation.  As we will demonstrate, for a simple 5-fold cross-validation, these differences are not large enough to detract from the effectiveness of the overall approach.  The second caveat is that this approach increases the training time required, roughly by the same factor as the number of folds used to create the cross-validated calibration scores.  While the idea of performing calibration in a cross-validated manner is not new (it is suggested in ~\cite{platt1999probabilistic}), it somehow escaped implementation in the current \texttt{sklearn} implementations of calibration.

To demonstrate this strategy, we again used the CIFAR-10 data set and the Convolutional Neural Net from the last example.  We divided the training set into 5 folds of 10K examples each and trained 5 different neural nets on the 5 different (40K-sized, overlapping) training sets for 400 epochs each and obtained predictions on the 5 (10K-sized, non-overlapping) "test" folds.  We concatenate the 5 10K-sized prediction matrices to obtain a 50K-sized set of cross-validated predictions for the entire training set.  We then compute a multi-class calibration function using Algorithm~\ref{alg3}.  Finally, we trained the same neural network model on the full 50K-sized training set.  We then used this final neural network to predict results on the (held-out) test set and computed the log-loss and accuracy both with and without applying the calibration function.  The results are in Figure~\ref{CIFAR_10_1} (c) and (d).  After 800 epochs, the uncalibrated log-loss was 0.3704, with clipping it was 0.3586 and with SplineCalib calibration it was 0.3286.  The uncalibrated accuracy was 88.86\% while the calibrated accuracy was 89.04\%.

\begin{algorithm}[t]
\caption{Cross-Validated SplineCalib}\label{alg4}
\begin{algorithmic}[0]
\State {\bf Input:} A feature matrix ${\vec{X}}$ and a corresponding vector $\vec{y}$  of actual $[1, \ldots, m]$ class values and a model to be trained. (We assume the model outputs probabilistic predictions)
\State {\bf Output:} A trained model which outputs (well-calibrated) probabilistic predictions.\\
\State (1) Train the model on the entire data set ${\vec{X}}$, $\vec{y}$.
\State (2) Split data into k folds 
\State (3) For each $i$ in 1 to k, designate fold $i$ as the test fold.  Train the model on the remaining folds and predict on fold $i$ to get matrix $\vec{Y_i}$ of predictions.  Matrix will be of size $n_i \times m$ where $n_i$ is the number of data points in fold $i$. 

\State (4) Stack the matrices $\vec{Y_i}$ to obtain a $n \times m$ matrix $\hat{\vec{Y}}$ of predicted probabilities. 
\State (5) Use Algorithm ~\ref{alg3} on $\hat{\vec{Y}}$ and $\vec{y}$ to obtain a calibration function $f: \mathcal{S}^m \rightarrow \mathcal{S}^m$
\State (6) Return a final model which composes the prediction of the model in Step (1) with the calibration function in step (5).

\end{algorithmic}
\end{algorithm}

This method allows major improvement in log-loss as well as substantial improvement in accuracy, without the need to "lose" data by setting aside a calibration set.  It is reasonable to suspect that many, if not most, deep learning models could be improved by this approach.  Of course, this is at a cost of considerably more computation time (roughly 5x for a 5-fold approach).  It is also notable that the log-loss for the calibrated version that was trained on 40K examples and calibrated on 10K examples (from the previous section) outperformed the uncalibrated version trained on 50K examples.  This suggests that, at least in some cases, it will be "worth it" to set aside data for calibration.  The improvement in accuracy, while small, is significant and consistent through the epochs of training the model.

\section{Implementation in Python}
The implementation of the SplineCalib algorithm used for the work in this paper is available in the \texttt{ml-insights} package.  It can be easily installed via \texttt{pip install ml\_insights}.  The code and several examples of its usage in tutorial Jupyter notebooks are available on github at \url{https://github.com/numeristical/introspective}.

\section{Summary and Future Work}
We presented {\em SplineCalib}, a new approach to calibration that improves on previous methods in several respects.  It uses cubic splines rather than piecewise constant functions or sigmoid functions.  It transforms the data to provide appropriate scaling for "over-confident" models. It handles multi-class problems elegantly, and can be used in a cross-validated scheme to efficiently use data.  It is broadly applicable and could improve the performance of many models, including those using deep learning.

This paper focused on log-loss.  If the Brier score is the metric of interest, one can use a similar approach where $L_2$-regularized {\em linear} regression (i.e. Ridge Regression) is substituted for the $L_2$-regularized Logistic regression.  In this case, some care must be take to avoid probabilities outside the interval [0,1]. 

There are several avenues for further research suggested here.  It is unknown how to choose the right amount of data for calibration.  There are several nuisance parameters (such as the $\epsilon$ in the compact logit function) that could be chosen more carefully.  Finally, there may be more efficient and clever ways to obtain a calibration set than the brute-force cross-validation approach, which would greatly improve the efficiency of the method.

\bibliography{nips_2017_spline_calib}

\end{document}